\documentclass[letterpaper]{article} 
\usepackage{aaai2026}  
\usepackage{times}  
\usepackage{helvet}  
\usepackage{courier}  
\usepackage[hyphens]{url}  
\usepackage{graphicx} 
\usepackage[numbers,square,comma,sort&compress]{natbib}  

\usepackage{caption} 
\usepackage{xcolor}
\usepackage{booktabs}
\usepackage{multirow}
\usepackage{amsmath}
\usepackage{amssymb}
\usepackage{algorithm}
\usepackage{algorithmic}
\usepackage{makecell}

\usepackage{newfloat}
\usepackage{listings}
\DeclareCaptionStyle{ruled}{labelfont=normalfont,labelsep=colon,strut=off} 
\floatstyle{ruled}
\newfloat{listing}{tb}{lst}{}
\floatname{listing}{Listing}
\title{HABIT: Chrono-Synergia Robust Progressive Learning Framework for Composed Image Retrieval}
\author{
    Zixu Li\textsuperscript{\rm 1},
    Yupeng Hu\textsuperscript{\rm 1}\thanks{Corresponding author.},
    Zhiwei Chen\textsuperscript{\rm 1},
    Shiqi Zhang\textsuperscript{\rm 1},
    Qinlei Huang\textsuperscript{\rm 1},
    Zhiheng Fu\textsuperscript{\rm 1},
    Yinwei Wei\textsuperscript{\rm 1}
}
\affiliations{
    \textsuperscript{\rm 1}School of Software, Shandong University\\
    \{lizixu.cs, zivczw, fuzhiheng8\}@gmail.com, 
    \{zhangshiqi, hql\}@mail.sdu.edu.cn, \\
    huyupeng@sdu.edu.cn,\
    weiyinwei@hotmail.com
}

\usepackage{bibentry}

\begin{document}
\maketitle

\begin{abstract}
Composed Image Retrieval (CIR) is a flexible image retrieval paradigm that enables users to accurately locate the target image through a multimodal query composed of a reference image and modification text. Although this task has demonstrated promising applications in personalized search and recommendation systems, it encounters a severe challenge in practical scenarios known as the Noise Triplet Correspondence (NTC) problem. This issue primarily arises from the high cost and subjectivity involved in annotating triplet data. To address this problem, we identify two central challenges: the \textbf{precise estimation of composed semantic discrepancy} and the \textbf{insufficient progressive adaptation to modification discrepancy}. To tackle these challenges, we propose a c\textbf{H}rono-synergi\textbf{A} ro\textbf{B}ust progress\textbf{I}ve learning framework for composed image re\textbf{T}rieval (\textbf{HABIT}), which consists of two core modules. First, the \textit{Mutual Knowledge Estimation} Module quantifies sample cleanliness by calculating the Transition Rate of mutual information between the composed feature and the target image, thereby effectively identifying clean samples that align with the intended modification semantics. Second, the \textit{Dual-consistency Progressive Learning} Module introduces a collaborative mechanism between the historical and current models, simulating human habit formation to retain good habits and calibrate bad habits, ultimately enabling robust learning under the presence of NTC. Extensive experiments conducted on two standard CIR datasets demonstrate that HABIT significantly outperforms most methods under various noise ratios, exhibiting superior robustness and retrieval performance.
Codes are available at https://github.com/Lee-zixu/HABIT
\end{abstract}

\begin{figure}[t]
\begin{center}
\includegraphics[width=\linewidth]{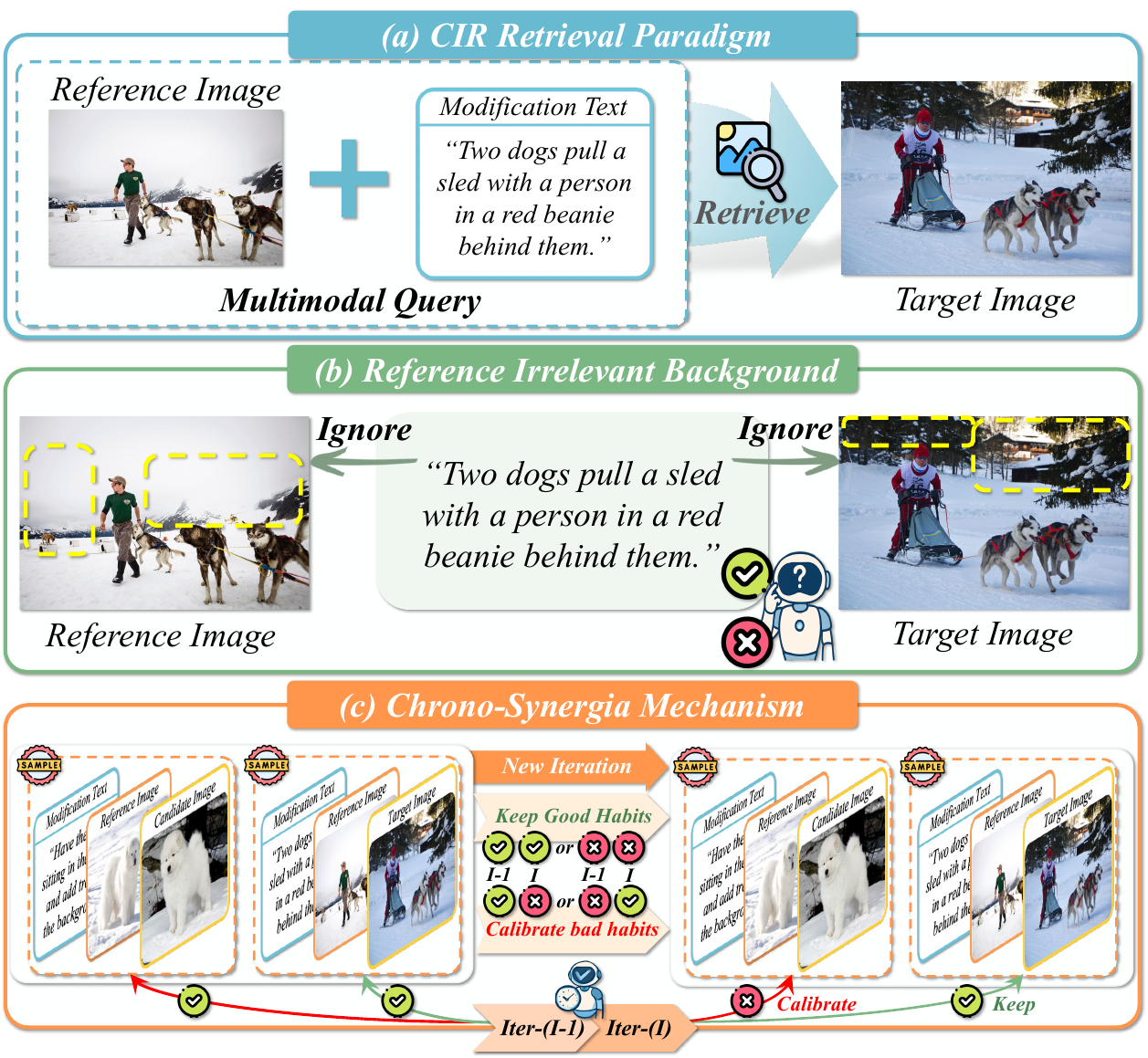}
\end{center}
   \caption{(a) presents an example of the CIR paradigm. (b) illustrates the commonly observed ``unmentioned visual discrepancies" in CIR task, which increase the difficulty of identifying Noise Triplet Correspondence. (c) depicts our proposed Chrono-Synergia Mechanism.}
\label{fig:intro}
\end{figure}
\section{Introduction}\label{sec:intro}
With the rapid growth of image data, Composed Image Retrieval (CIR)~\cite{limn, sprc, INTENT,REFINE,ReTrack} has emerged as a key research focus in information retrieval~\cite{LiuIMP-GCN,ssn, cala,cheng2026enhancing,pushe,xie2026conquer,xie2026hvd,targetrecognition,MAIL} and multimodal learning~\cite{wang2024twin,chen2025autoneural,jia2026ramrecover3dhuman,li2025human,li2026multiple,ni2025wonderturbo}. As shown in Figure~\ref{fig:intro}(a), CIR differs from unimodal retrieval by supporting multimodal queries consisting of a reference image and modification text, which enables the retrieval of a semantically consistent target image. Although significant progress has been made in the CIR task, real-world applications~\cite{qiu2025duet,zhao2024balf,zhang2026decoding,liu2024dvlo,jiang2025stg,yuan2025deep,lu2023tf,anonymous2025comptrack,zhou2025dragflow,lan2025multi,yu2025cotextor,liao2025convex,qiu2024tfb,li2025set,duan2025copinn,focustrack} still face substantial challenges. Due to the high cost and inherent subjectivity of triplet annotation, practical datasets are often affected by annotation errors and inaccurate semantic alignment~\cite{xie2025chat,liu2024unsupervised,xie2026delving,gu2025mocount,li2025chatmotion,jia2024adaptive,liu2024graph,liu2023retrieval,STABLE, ERASE}. Moreover, this issue is further exacerbated in large-scale datasets involving large models, owing to the hallucination problem commonly observed in such models. To mitigate this problem, Li et al.~\cite{TME} introduced the Noise Triplet Correspondence (NTC) to enhance the robustness of CIR models.

Unlike traditional cross-modal retrieval tasks such as video-text matching~\cite{liu2018attentive,hu2021video,liu2018cross,hu2023semantic,LiuSALLM,liu2025queries}, CIR~\cite{tgcir, encoder, HUD, OFFSET} is inherently a semantic modification task where the reference image and modification text often contain inconsistent semantics, increasing the complexity of the NTC problem. As a result, existing robust methods for cross-modal matching~\cite{matching-NDC-1, matching-NDC-2} fail to generalize to CIR. Moreover, during the CIR annotation process~\cite{FashionIQ, shoes}, annotators receive a reference and target image pair and are asked to describe their differences as modification text. Due to annotation cost constraints, other candidate images are not shown, leading to modification texts that are often brief and omit subtle differences. As shown in Figure~\ref{fig:intro}(b), visual details such as ``sky'' and ``mountains'' in the reference image as well as ``tree'' and ``house'' in the target image are frequently excluded, resulting in unmentioned visual discrepancies. These discrepancies create persistent semantic gaps between the composed feature and the target image, which vary across samples. However, current robust CIR methods such as TME~\cite{TME} rely heavily on feature similarity to identify noisy correspondences, which may misclassify samples with diverse semantic gaps and ultimately limit model performance and generalization.

However, addressing the above limitations remains challenging due to two key factors.
\textbf{(1) Precise estimation of composed semantic discrepancy.} To identify unmentioned visual discrepancies and circumvent the limitations of similarity-based noise assessment methods, it is essential to develop an approach capable of accurately quantifying the fine-grained semantic consistency between the composed feature and the target image. Moreover, the semantic fusion of the reference image and modification text in CIR is inherently complex, making it difficult to directly apply conventional robust learning methods. Thus, developing an effective estimation mechanism for composed semantic discrepancy constitutes the first major challenge.
\textbf{(2) Insufficient progressive adaptation to modification discrepancy.} Due to the intrinsic semantic gap between the reference image and modification text, CIR models often misinterpret this discrepancy during early training stages, leading to clean positive samples being wrongly treated as noisy correspondences. Therefore, the second challenge lies in designing a learning strategy that progressively enhances the model's ability to adapt to diverse triplet compositions and continuously reduces the risk of misdetermination.

To address the aforementioned challenges, we propose a c\textbf{H}rono-synergi\textbf{A} ro\textbf{B}ust progress\textbf{I}ve learning framework for composed image re\textbf{T}rieval (\textbf{HABIT}). HABIT tackles the NTC problem in CIR by modeling mutual knowledge and leveraging progressive learning via dual knowledge collaboration. It comprises two key modules:
(1) \textit{Mutual Knowledge Estimation} Module quantifies sample cleanliness by computing the Transition Rate of mutual knowledge between the composed feature and the target image. This facilitates the reliable identification of clean samples that align with the intended modification semantics, improving noise recognition accuracy.
(2) \textit{Dual-consistency Progressive Learning} addresses insufficient adaptation to modification discrepancy. As shown in Figure~\ref{fig:intro}(c), we introduce the Chrono-Synergia Mechanism, which simulates human habit formation by integrating predictions from both historical and current models. Historically consistent decisions are retained as good habits, while inconsistent ones are calibrated as bad habits, enabling robust learning in the presence of NTC.

In summary, the contributions of this paper are threefold:
\begin{itemize}
	\item We conduct an in-depth analysis of the NTC problem in CIR, and for the first time, we identify two key challenges faced by existing methods: the precise estimation of composed semantic discrepancy and the insufficient progressive adaptation to modification discrepancy.
	\item We propose a novel robust learning framework for CIR, named HABIT, which employs the Transition Rate of mutual knowledge for accurate noise-aware label assignment. In addition, HABIT simulates human habit formation and achieves robust learning under noisy triplet conditions through Dual-consistency Progressive Learning.
	\item Extensive experiments on two standard CIR benchmarks demonstrate that HABIT outperforms most methods across varying noise levels, significantly improving retrieval performance in noisy settings.
\end{itemize}

\begin{figure*}
\begin{center}
\includegraphics[width=\linewidth]{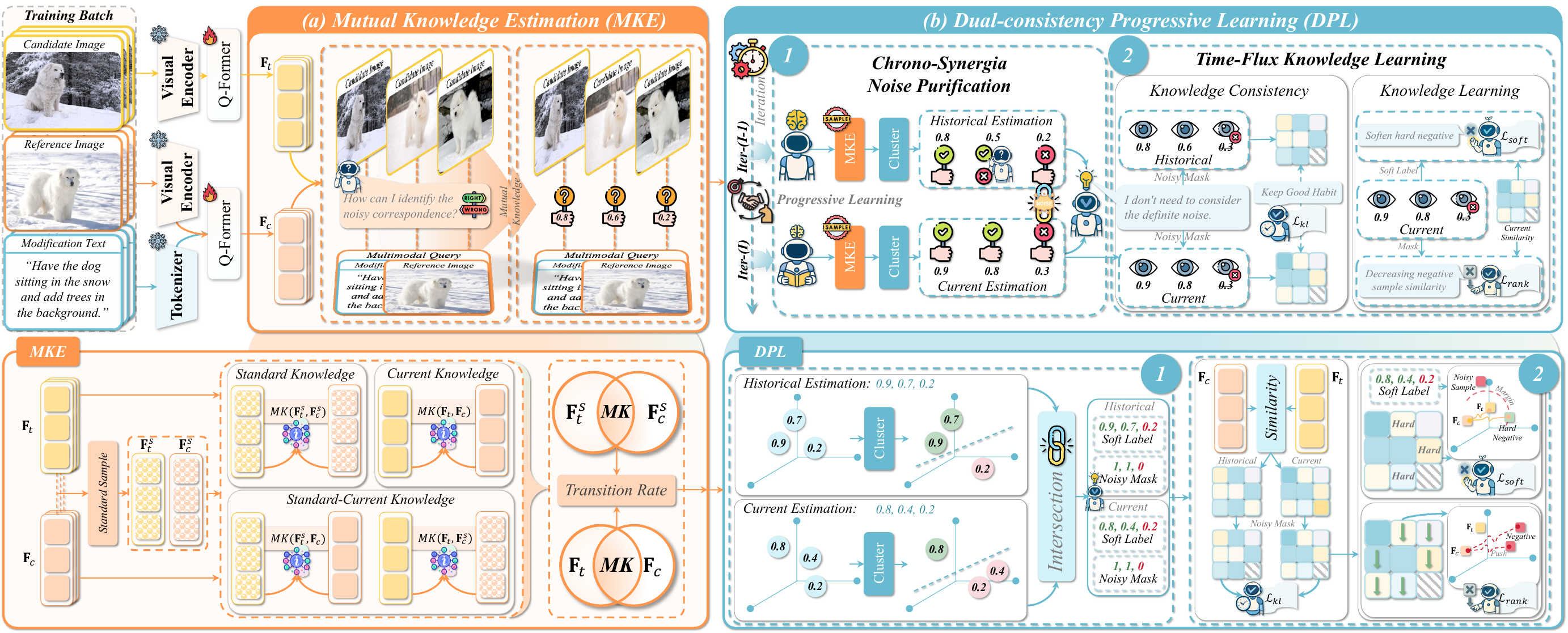}

\end{center}
   \caption{HABIT consists of two modules: (a) Mutual Knowledge Estimation and (b) Dual-consistency Progressive Learning. }
\label{fig:overview}
\end{figure*}

\section{Related Work}

\subsection{Conventional Composed Image Retrieval}
Composed Image Retrieval (CIR) plays a crucial role in computer vision~\cite{song6,pillarhist,liu2024difflow3d,liu2025difflow3d,jiang2026foeforesterrorsmakes,gao2024eraseanything,song2,yu2025visualizing,liao2024globalpointer,zhang2023multi,bi2025llava,li2023hong,zhou2024lidarptq,yu2025yielding,zhou2023fastpillars,wang2026eeo,wang2024computing,liu2023regformer,Feng2023MaskCon,wen2023syreanet,yu2025iidm,song1} and multimodal learning~\cite{wang2025ascd,yu2025visual,song4,zhang2024cf,dong2026neureasonerexplainablecontrollableunified,Feng2024CLIPCleaner,song5,zhang2026towards,bi2025cot,bi2025prismselfpruningintrinsicselection,jiang2025transforming,song3,meng2026tri,zhou2024information,song13,dong2025aurora}, aiming to retrieve the target image based on a reference image and modification text. Existing approaches can generally be categorized into two types. The first category consists of models that use traditional architectures (e.g., ResNet, LSTM) to extract image and text features, followed by the multimodal composition.
In contrast, the second group leverages Vision-Language Pretrained (VLP) models such as CLIP~\cite{clip}, BLIP-2~\cite{blip-2}, to extract multimodal query features, then apply relatively simple strategies for feature alignment and composition to achieve remarkable performance~\cite{cala,sprc,HINT, MELT}. Although some methods have investigated false positive samples~\cite{mgur}, their primary objective is to improve the performance of traditional CIR, making them less effective from the perspective of NTC.

\subsection{Noisy Correspondence Learning}
While robust learning~\cite{he2024robust,pu2025robust,lu2024robust,sun2025roll,Feng2022SSR,Feng2024NoiseBox,xu2025noisy,BML,sun2024robust} has been widely explored in multimodal tasks~\cite{noisy-label-1,TME,ye2026gigaworld,ni2025recondreamerRL,ni2025recondreamer}, recent work increasingly focuses on the more complex issue of noisy correspondence across modalities. This type of noise, which surpasses simple label errors, often induces overfitting and performance degradation. Existing approaches in visual-language pretraining and cross-modal learning~\cite{noisy-vlm, TME} mainly consider two-modality noise, overlooking the challenges in real-world Composed Image Retrieval. To address this, Li et al.~\cite{TME} introduced the Noise Triplet Correspondence (NTC) problem, underscoring semantic ambiguity and annotation errors in large-scale datasets. While prior alignment-based strategies~\cite{TME} have enhanced robustness, our HABIT further advances this by leveraging fine-grained mutual knowledge modeling and Dual-consistency Progressive Learning, yielding substantial gains in robustness to NTC.

\section{HABIT}
As a primary innovation, the proposed HABIT is designed to quantify sample cleanliness via the mutual knowledge transition  rate and to achieve robust learning in the \textit{NTC} environment by simulating human habit formation through a dual-consistency progressive learning strategy. As shown in Figure~\ref{fig:overview}, HABIT comprises two core modules:
\textit{(a) Mutual Knowledge Estimation (MKE)} and \textit{(b) Dual-consistency Progressive Learning (DPL)}.
In this section, we first formulate CIR task with NTC and then elaborate on each module.

\subsection{Problem Formulation}
The CIR task seeks to retrieve the target image that matches a given multimodal query from an image database. In practice, CIR datasets frequently suffer from annotation errors within their triplets, termed NTC, with such erroneous samples referred to as noisy triplets. Noisy triplets typically fall into two types: (1) partial match: the modification text $x_m$ partially describes the transformation from the reference image $x_r$ to the target image $x_t$; (2) full mismatch: $x_m$ completely misrepresents the modification. Following TME~\cite{TME}, we simulate noisy scenarios by randomly selecting a subset of training triplets according to a noise ratio $\sigma$.
Given a triplet set with NTC, $\mathcal{T} = \bigl\{ \langle x_r, x_m, x_t \rangle_n \bigr\}_{n=1}^{N},$ where $x_r$, $x_m$, and $x_t$ may not be properly aligned, the objective is to learn an embedding function $\mathcal{G}$ that maps the multimodal query $(x_r, x_m)$ close to its corresponding target image $x_t$ in a shared metric space, as $\mathcal{G}(x_r, x_m)\rightarrow\mathcal{G}(x_t),$
where $\mathcal{G}$ is the embedding function to be learned, mapping multimodal queries and target images into one metric space.

\subsection{Mutual Knowledge Estimation (MKE)}
This module quantifies sample cleanliness by measuring the mutual knowledge transition rate between the composed feature and the target image, enabling effective identification of samples whose semantics align with the modification text and improving noisy correspondence detection. First, the module extracts the composed feature and modality features from the target image. Next, we compute the semantic matching degree between the composed feature and the target feature based on mutual knowledge across modalities, and the mutual knowledge transition rate is defined to capture semantic discrepancies. This metric serves as the criterion for evaluating the semantic noise margin.

Specifically, we employ Q-Former, which is proven effective in various CIR models~\cite{sprc,covr-2}, to extract cross-modal features from both visual and textual modalities, extracting precise composed and target features, as formulated below,
\begin{equation}
\mathbf{F}_c\!\!=\!\!\operatorname{Q-Former}(\varPhi_{\mathbb{I}}(x_r), \varPhi_{\mathbb{T}}(x_m)),\mathbf{F}_t\!\!=\!\!\operatorname{Q-Former}(\varPhi_{\mathbb{I}}(x_t)), 
\end{equation}
where $\mathbf{F}_c, \mathbf{F}_t \in \mathbb{R}^{Q \times D}$ denote the composed feature and the target feature, respectively, $Q$ is the number of learnable queries, $D$ is the embedding dimension, and $\varPhi_{\mathbb{I}}$ and $\varPhi_{\mathbb{T}}$ denote the visual encoder and the text tokenizer, respectively.

Subsequently, we define the mutual knowledge between the composed feature $\mathbf{F}_c$ and the target feature $\mathbf{F}_t$ for any sample in the batch, which is used to measure the semantic matching degree between the two features, as formulated as,
\begin{equation}
\operatorname{MK}(\mathbf{F}_c, \mathbf{F}_t) = \sum_{\mathbf{f}_c \in \mathbf{F}_c} \sum_{\mathbf{f}_t \in \mathbf{F}_t} p(\mathbf{f}_c, \mathbf{f}_t) \log \frac{p(\mathbf{f}_c,\mathbf{f}_t)}{p(\mathbf{f}_c)p(\mathbf{f}_t)},
\end{equation}
where $p(\mathbf{f}_c, \mathbf{f}_t)$ denotes the joint probability distribution between $\mathbf{f}_c$ and $\mathbf{f}_t$, and $p(\mathbf{f}_c)$ and $p(\mathbf{f}_t)$ denote the marginal probability distributions of $\mathbf{f}_c$ and $\mathbf{f}_t$, respectively.

However, the mutual knowledge value alone does not provide a clear margin between noisy and correctly matched samples. To address this, inspired by~\cite{TSVC}, in each batch, we select the triplet with the lowest loss as the \textit{Standard Sample} (${\mathbf{F}_c^s,\mathbf{F}_t^s}$), assumed to be the cleanest correspondence. We then calculate the mutual knowledge discrepancy of all other samples relative to this standard sample to define the mutual knowledge transition rate, which quantifies each sample's discrepancy from the standard and serves as the estimation of noisy correspondence, formulated as,
\begin{equation}
\operatorname{TR}(\mathbf{F}_{c}, \mathbf{F}_{t}) = \frac{|\operatorname{MK}(\mathbf{F}_{c}^{s}, \mathbf{F}_{t}^{s}) - \operatorname{MK}(\mathbf{F}_{c}, \mathbf{F}_{t})|}{\operatorname{MK}(\mathbf{F}_{c}^{s}, \mathbf{F}_{t}^{s})}.
\label{TR}
\end{equation}

Furthermore, since noisy correspondence can originate from either the composed feature (i.e., reference image/modification text side) or the target feature, we also consider the transition rates between each composed/target feature and the corresponding standard sample, denoted as $\operatorname{TR}(\mathbf{F}_{c}, \mathbf{F}_{t}^s)$ and $\operatorname{TR}(\mathbf{F}_{t}, \mathbf{F}_{c}^s)$. Using these transition rates, we estimate the cleanliness of each triplet sample as follows,
\begin{equation}
\mathbb{E}\!=\!{(1 + \operatorname{TR}(\mathbf{F}_{c}, \mathbf{F}_{t})\!+\!|\operatorname{TR}(\mathbf{F}_{c}, \mathbf{F}_{t}^s)\!-\!\operatorname{TR}(\mathbf{F}_{t}, \mathbf{F}_{c}^s)|)}^{-\!1}\!.
\label{estimation}
\end{equation}

To be specific, when the transition rate between a sample's composed feature and its target feature is small, and when both $\operatorname{TR}(\mathbf{F}_{c}, \mathbf{F}_{t}^s)$ and $\operatorname{TR}(\mathbf{F}_{t}, \mathbf{F}_{c}^s)$ are close to that of the standard sample, it indicates that the matching relationship between the multimodal query composed feature and the target image is more reliable, i.e. the sample is more likely a clean correspondence.

\subsection{Dual-consistency Progressive Learning (DPL)}
The \textit{Dual-consistency Progressive Learning (DPL)} module is designed to address the limited progressive adaptivity caused by modification discrepancies, allowing the model to incrementally adapt to the challenges of noisy correspondence detection and enhance robustness. Building on sample cleanliness estimations from the MKE module, DPL incorporates the \textit{Chrono-Synergia Noise Discrimination} and \textit{Time-Flux Knowledge Updating}. These enable HABIT to simulate human habit formation by integrating predictions from both the historical and current models, preserving good habits and calibrating bad habits for robust learning in NTC scenarios. We now explain this module in detail.

\begin{table*}[ht!]
\centering
\begin{tabular}{c|l|cc|cc|cc|cc|ccc}
\hline
\multirow{2}{*}{Noise} & \multirow{2}{*}{Methods} 
& \multicolumn{2}{c|}{Dress} 
& \multicolumn{2}{c|}{Shirt} 
& \multicolumn{2}{c|}{Toptee} 
& \multicolumn{3}{c}{Average} \\
\cline{3-11}
& & R@10 & R@50 & R@10 & R@50 & R@10 & R@50 & R@10 & R@50 & AVG. \\

\hline
\multirow{8}{*}{0\%}

& SSN~\cite{ssn}~(AAAI'24) & 34.36 & 60.78 & 38.13 & 61.83 & 44.26 & 69.05 & 38.92 & 63.89 & 51.40 \\
& CALA~\cite{cala}~(SIGIR'24) & 42.38 & 66.08 & 46.76 & 68.16 & 50.93 & 73.42 & 46.69 & 69.22 & 57.96 \\
& SPRC~\cite{sprc}~(ICLR'24) & 49.18 & \underline{72.43} & 55.64 & 73.89 & \underline{59.35} & \underline{78.58} & 54.92 & 74.97 & 64.85 \\
& RCL~\cite{RCL}~(TPAMI'23) & 48.79 & \textbf{72.68} & 55.89 & 73.90 & 56.91 & 77.41 & 53.86 & 74.66 & 64.26 \\
& RDE~\cite{RDE}~(CVPR'24) & 47.84 & 71.89 & 54.37 & 73.55 & 56.91 & 77.21 & 53.04 & 74.22 & 63.63 \\
& TME~\cite{TME}~(CVPR'25) & \underline{49.73} & 71.69 & \underline{56.43} & \underline{74.44} & 59.31 & \textbf{78.94} & \underline{55.15} & \underline{75.02} & \underline{65.09} \\
& \textbf{HABIT (Ours)} & \textbf{49.99} & 72.38 & \textbf{56.62} & \textbf{74.68} & \textbf{59.51} & 78.53 & \textbf{55.38} & \textbf{75.20} & \textbf{65.29}\\

\hline
\multirow{8}{*}{20\%}
& SSN~\cite{ssn}(AAAI'24) & 22.61 & 45.56 & 27.87 & 48.58 & 31.82 & 55.28 & 27.43 & 49.81 & 38.62 \\
& CALA~\cite{cala}(SIGIR'24) & 29.05 & 51.36 & 35.28 & 56.23 & 36.05 & 58.24 & 33.46 & 55.28 & 44.37 \\
& SPRC~\cite{sprc}(ICLR'24) & 39.81 & 62.22 & 48.58 & 66.29 & 50.48 & 70.58 & 46.29 & 66.36 & 56.33 \\
& RCL~\cite{RCL}(TPAMI'23) & 47.05 & \underline{70.65} & 53.14 & 71.74 & 55.28 & 75.62 & 51.82 & 72.67 & 62.25 \\
& RDE~\cite{RDE}(CVPR'24) & 44.62 & 68.91 & 50.74 & 69.09 & 52.12 & 73.38 & 49.16 & 70.64 & 59.81 \\
& TME~\cite{TME}(CVPR'25)& \underline{49.03} & 70.35 & \textbf{55.84} & \underline{73.16} & \underline{57.22} & \underline{78.23} & \underline{54.03} & \underline{73.91} & \underline{63.97} \\
& \textbf{HABIT (Ours)} & \textbf{49.63 }& \textbf{71.34} & \underline{55.67}& \textbf{73.19 }& \textbf{58.14} & \textbf{78.32} & \textbf{54.48} & \textbf{74.28} & \textbf{64.38} \\
\hline
\multirow{8}{*}{50\%}
& SSN~\cite{ssn}~(AAAI'24) & 15.27 & 33.71 & 23.36 & 41.61 & 22.79 & 42.94 & 20.47 & 39.42 & 29.95 \\
& CALA~\cite{cala}~(SIGIR'24) & 20.77 & 40.95 & 29.69 & 46.57 & 27.03 & 46.81 & 24.83 & 44.78 & 34.80 \\
& SPRC~\cite{sprc}~(ICLR'24) & 35.94 & 57.16 & 42.25 & 61.63 & 44.98 & 64.76 & 41.06 & 61.19 & 51.12 \\
& RCL~\cite{RCL}~(TPAMI'23) & 43.68 & 66.44 & 50.74 & 69.19 & 52.63 & 73.84 & 49.01 & 69.82 & 59.42 \\
& RDE~\cite{RDE}~(CVPR'24) & 41.30 & 64.75 & 47.06 & 66.34 & 50.13 & 70.63 & 46.16 & 67.24 & 56.70 \\
& TME~\cite{TME}~(CVPR'25)& \underline{46.26} & \underline{68.27} & \underline{53.09} & \underline{71.88} & \underline{55.07} & \underline{76.59} & \underline{51.47} & \underline{72.25} & \underline{61.86} \\
& \textbf{HABIT (Ours)} & \textbf{47.33} & \textbf{69.71} & \textbf{53.72} & \textbf{72.55} & \textbf{56.51} & \textbf{77.00} & \textbf{52.52} & \textbf{73.09} & \textbf{62.80} \\
\hline
\multirow{8}{*}{80\%}
& SSN~\cite{ssn}~(AAAI'24) & 11.16 & 25.24 & 16.98 & 30.72 & 17.03 & 32.64 & 15.05 & 29.53 & 22.29 \\
& CALA~\cite{cala}~(SIGIR'24) & 14.28 & 30.59 & 19.73 & 35.82 & 19.48 & 36.10 & 17.83 & 34.41 & 26.00 \\
& SPRC~\cite{sprc}~(ICLR'24) & 28.41 & 50.77 & 36.21 & 54.37 & 35.90 & 59.06 & 33.51 & 54.03 & 43.77 \\
& RCL~\cite{RCL}~(TPAMI'23) & 38.82 & 60.54 & 45.44 & 64.38 & 47.42 & 68.38 & 43.89 & 64.43 & 54.16 \\
& RDE~\cite{RDE}~(CVPR'24) & 37.63 & 59.64 & 43.62 & 62.12 & 46.10 & 66.50 & 42.45 & 62.75 & 52.60 \\
& TME~\cite{TME}~(CVPR'25)& \underline{41.45} & \underline{64.35} & \underline{47.30} & \underline{68.20} & \underline{51.25} & \underline{73.23} & \underline{46.67} & \underline{68.60} & \underline{57.63} \\
& \textbf{HABIT (Ours)} & \textbf{42.04} & \textbf{65.20} & \textbf{50.12} & \textbf{69.77} & \textbf{52.92} & \textbf{73.61} & \textbf{48.36} & \textbf{69.53} & \textbf{58.94} \\
\hline
\end{tabular}
\caption{Performance comparison on FashionIQ in terms of R@K (\%). The best result under each noise ratio is highlighted in \textbf{bold}, while the second-best result is \underline{underlined}.}
\label{tab:fiq}
\end{table*}

\noindent\textbf{Chrono-Synergia Noise Discriminantion}
Intuitively, the cleanliness estimations of standard samples remain relatively stable during training, whereas those of noisy correspondence samples tend to fluctuate and deviate from clean samples. Based on this, we introduce the \textit{Chrono-Synergia Noise Discrimination}, which aims to chrono-synergize the dynamic changes of cleanliness estimations to identify noisy triplets that fall outside the normal matching range.

Specifically, we denote $\mathbf{e}^{(I)}\!\!=\!\![\mathbb{E}_1^{(I)}, \dots, \mathbb{E}_B^{(I)}]$ as the estimation sequence for all samples in the batch at the $I$-th iteration, where $B$ is the batch size. For each iteration's estimation sequence, we apply the DBSCAN~\cite{dbscan} to obtain the set of outlier samples $\mathcal{O}^{(I)}$ for that iteration. Similarly, by applying the same procedure to the historical estimation sequence from the previous iteration $(I\!\!-\!\!1)$, we obtain the corresponding outlier set $\mathcal{O}^{(I-1)}$.

Subsequently, we merge the outlier sets from both the historical and current iterations to compute the noisy mask $\mathbf{m}^{(I)}=[m_1,\dots,m_b,\dots,m_B]$ for the current batch at iteration $I$, where the noisy mask $m_b$ for the $b$-th triplet sample in the batch, formalized as follows,
\begin{equation}
m_b =
\begin{cases}
$0$, & \text{if } \{b\} \in \mathcal{O}^{(I)} \cap \mathcal{O}^{(I-1)} \\
$1$, & \text{otherwise}
\end{cases}.
\label{noisymask}
\end{equation}

The resulting noisy mask can accurately identify samples that are outliers at both time points, synergizing the discrimination capabilities of the historical and current models for noisy triplets, and thereby continuously filtering out those noisy triplets that are stably recognized.

\noindent\textbf{Time-Flux Knowledge Updating}
Since the noise discrimination mechanism applies strict criteria, some mismatched samples may still be misclassified, leading to model's ``bad habits''. To further improve robustness against residual noisy correspondence and calibrate these bad habits while keeping good ones, we introduce \textit{Time-Flux Knowledge Updating}, which utilizes the temporal evolution of cleanliness estimations to maintain semantic consistency and clear time-flux margins, thus enhancing training robustness.

\textbf{\textit{Knowledge Consistency.} }
Before calibrating model's bad habits, it is essential to preserve its learned good habits. Thus, to ensure semantic consistency and maintain accurate matching capabilities throughout training, we introduce \textit{Knowledge Consistency}. It constrains the evolution of the model's triplet semantic matching degree across iterations. By comparing the similarity distributions produced at the current iteration $I$ and the previous iteration $(I-1)$, this mechanism prevents the model from losing its ``good habits'' (the correctly learned determination capabilities).

Specifically, let $I$ denote the current iteration. Define
$\mathbf{s}^{(I)}\in\mathbb{R}^{B\times B}$
as the similarity matrix between all multimodal query features $\mathbf{F}_c$ and all target image features $\mathbf{F}_t$ in the current batch, and let $\mathbf{s}^{(I-1)}$ denote the corresponding similarity matrix from the previous iteration on the same batch. Correspondingly, $\mathbf{m}^{(I)}$ and $\mathbf{m}^{(I-1)}$ represent the current and previous noisy masks, respectively. We enforce a consistency constraint on ``good habits'' by minimizing the Kullback-Leibler (KL) divergence between the similarity distributions of consecutive iterations, as follows,
\begin{equation}
\mathcal{L}_{KL} = \frac{1}{B} \sum_{b=1}^{B} D_{KL} \left( \left(\mathbf{s}^{(I)}_b \cdot \textbf{m}_b^{(I)} \right) \Vert \left( \mathbf{s}^{(I-1)}_b \cdot \textbf{m}_b^{(I-1)} \right)\right),
\label{kl}
\end{equation}
where $\mathbf{s}^{(I)}_b, \mathbf{s}^{(I-1)}_b$ denote the similarity vectors of the $b$-th multimodal query within the batch at iterations $I$ and $I-1$.

\textbf{\textit{Knowledge Learning.}}
While preserving good habits, we also need to calibrate bad habits formed during previous training to mitigate the harm caused by noisy correspondence misdetermination. To this end, we introduce the soft estimation margin loss $\mathcal{L}_{soft}$ and the robust contrastive loss $\mathcal{L}_{rank}$, performing robust optimization on negative examples with higher similarity to the current sample, thereby jointly enhancing the model's discriminative capability and robustness to noise interference.

To address model misdetermination, we adopt a margin softening strategy based on cleanliness estimations. Let $I$ denote the current iteration, and $\textbf{s}^{(I)} \in \mathbb{R}^{B\times B}$ represent the similarity matrix between all composed features $\textbf{F}_c$ and target images $\textbf{F}_t$ in the batch. The noise mask for iteration $I$ is denoted as $\textbf{m}^{(I)}$. We apply this mask to identify noisy correspondences and employ a margin-based hard negative sampling strategy~\cite{noisy-nips}. The soft estimation margin loss is then defined as follows:
\begin{equation}
\mathcal{L}_{soft}\!=\!\frac{1}{B} \sum_{b=1}^{B} \!\left(\max_{j\neq b}\left[\operatorname{margin}(\textbf{e}^{(I)}_b)\!+\!\textbf{s}_{bj}^{(I)}\!-\! \textbf{s}_{bb}^{(I)} \right]_+\!\!\!\cdot\! \textbf{m}_b^{(I)}\right),
\label{soft}
\end{equation}
where $[\cdot]+ = \max(0, \cdot)$ denotes the ReLU operation, and $\textbf{s}{bb}$ represents the similarity of the diagonal samples. $\operatorname{margin}(\textbf{e}^{(I)}_b)$ is a dynamic margin that depends on the estimation value $\textbf{e}^{(I)}_b$ of the $b$-th sample in the batch.

Subsequently, following~\cite{TME,RCL}, we utilize the robust contrastive loss, which actively increases the distance between the multi-modal query and negative samples by reducing their similarity, and prevents interference from noisy correspondence, formulated as,
\begin{equation}
    \mathcal{L}_{rank} \!=\! \frac{1}{B} \!\sum_{b=1}^{B} \!-\!\log\! \left(\left( 1-\operatorname{Softmax}(\textbf{s}^{(I)}_b/\tau) \right)\cdot \textbf{m}_b^{(I)}\right),
    \label{rank}
\end{equation}
where $B$ is the batch size, $\tau$ is the temperature coefficient, $\textbf{m}^{(I)}$ refers to the noisy mask of the $I$-th iteration batch, and $\textbf{s}^{(I)}\!\!\in\!\! \mathbb{R}^{B\times B}$ is the similarity matrix between all composed features $\textbf{F}_c$ and target images $\textbf{F}_t$ in the $I$-th iteration batch.

Finally, we obtain the final loss function of HABIT as,
\begin{equation}
    \mathbf{\Theta^{*}}=
    \underset{\mathbf{\Theta}}{\arg \min } \left( {\mathcal{L}}_{rank}+\kappa {\mathcal{L}}_{KL} +\gamma {\mathcal{L}}_{soft} \right),
    \label{optimization}
\end{equation}
where $\mathbf{\Theta^{*}}$ is the to-be-optimized parameter for HABIT and $\kappa, \gamma$ are the trade-off hyper-parameters.

\begin{table*}[t]
\centering
\setlength{\tabcolsep}{6pt}
\begin{tabular}{c|l|cccc|ccc|c}
\hline

\multirow{2}{*}{\textbf{Noise}} & \multirow{2}{*}{\textbf{Methods}} 
& \multicolumn{4}{c|}{R@K} 
& \multicolumn{3}{c|}{R$_{\text{sub}}$@K} 
& \multirow{2}{*}{Avg(R@5, R$_{\text{sub}}$@1)} \\
\cline{3-9}
& & K=1 & K=5 & K=10 & K=50 & K=1 & K=2 & K=3 & \\
\hline

\multirow{8}{*}{0\%}
& SSN~\cite{ssn}~(AAAI'24) & 43.91 & 77.25 & 86.48 & 97.45 & 71.76 & 88.63 & 95.54 & 74.51 \\
& CALA~\cite{cala}~(SIGIR'24) & 49.11 & 81.21 & 89.59 & 98.00 & 76.27 & 91.04 & 96.46 & 78.74 \\
& SPRC~\cite{sprc}~(ICLR'24) & 51.96 & 82.12 & 89.74 & 97.69 & 80.65 & 92.31 & 96.60 & 81.39 \\
& RCL~\cite{RCL}~(TPAMI'23) & 53.16 & 82.41 & 90.12 & \textbf{98.34} & 79.57 & 92.02 & 96.87 & 80.99 \\
& RDE~\cite{RDE}~(CVPR'24) & 51.81 & 82.02 & \underline{90.60} & 97.93 & 78.17 & 91.90 & 96.70 & 80.10 \\
& TME~\cite{TME}~(CVPR'25) & \textbf{53.42} & \textbf{82.99} & 90.24 & 98.15 & \textbf{81.04} & \underline{92.58} & \underline{96.94} & \textbf{82.01} \\
& \textbf{HABIT (Ours)} & \underline{52.71} & \underline{82.64} & \textbf{90.63} & \underline{98.19} & \underline{80.99} & \textbf{92.77} & \textbf{97.00} & \underline{81.82} \\

\hline
\multirow{7}{*}{20\%}
& SSN~\cite{ssn}~(AAAI'24) & 34.02 & 65.90 & 75.78 & 91.33 & 66.92 & 85.90 & 93.45 & 66.41 \\
& CALA~\cite{cala}~(SIGIR'24) & 41.33 & 72.70 & 82.84 & 94.34 & 71.66 & 88.15 & 94.94 & 72.18 \\
& SPRC~\cite{sprc}~(ICLR'24) & 45.90 & 75.86 & 83.52 & 93.37 & 78.10 & \underline{91.40} & 96.05 & 76.98 \\
& RCL~\cite{RCL}~(TPAMI'23) & 50.43 & \textbf{81.11} & \underline{88.82} & 96.68 & 77.52 & 90.80 & 95.71 & 79.31 \\
& RDE~\cite{RDE}~(CVPR'24) & 49.23 & 78.63 & 86.80 & 95.78 & 76.58 & 90.31 & 96.07 & 77.60 \\
& TME~\cite{TME}~(CVPR'25) & \underline{51.35} & 81.01 & 88.53 & \underline{97.81} & \textbf{78.46} & 91.25 & \underline{96.39} & \textbf{79.74} \\
& \textbf{HABIT (Ours)} & \textbf{51.68} & \underline{81.02} & \textbf{89.24} & \textbf{97.81} & \underline{78.20} & \textbf{91.66} & \textbf{96.75} & \underline{79.61} \\
\hline
\multirow{7}{*}{50\%}
& SSN~\cite{ssn}~(AAAI'24) & 25.93 & 53.71 & 63.40 & 82.10 & 62.10 & 82.27 & 91.57 & 57.90 \\
& CALA~\cite{cala}~(SIGIR'24) & 36.10 & 66.12 & 77.76 & 92.10 & 68.12 & 85.66 & 93.59 & 67.12 \\
& SPRC~\cite{sprc}~(ICLR'24) & 39.93 & 66.00 & 73.59 & 86.48 & 75.81 & 89.21 & 95.37 & 70.90 \\
& RCL~\cite{RCL}~(TPAMI'23) & \underline{48.58} & 77.45 & 85.93 & 94.70 & 75.60 & 89.28 & 94.80 & 76.52 \\
& RDE~\cite{RDE}~(CVPR'24) & 45.98 & 75.30 & 83.73 & 94.48 & 73.98 & 88.99 & 95.13 & 74.64 \\
& TME~\cite{TME}~(CVPR'25) & 48.48 & \underline{78.94} & \underline{87.28} & \underline{96.99} & \underline{76.48} & \underline{90.07} & \underline{95.83} & \underline{77.71} \\
& \textbf{HABIT (Ours)} & \textbf{50.32} & \textbf{79.63} & \textbf{88.34} & \textbf{97.06} & \textbf{76.84} & \textbf{90.60} & \textbf{96.27} & \textbf{78.87} \\
\hline
\multirow{8}{*}{80\%}
& SSN~\cite{ssn}~(AAAI'24) & 20.48 & 43.98 & 54.27 & 74.80 & 56.48 & 77.20 & 89.54 & 50.23 \\
& CALA~\cite{cala}~(SIGIR'24) & 31.52 & 61.49 & 72.60 & 89.86 & 64.34 & 83.52 & 92.60 & 62.92 \\
& SPRC~\cite{sprc}~(ICLR'24) & 29.95 & 51.25 & 58.51 & 73.86 & 70.22 & 86.05 & 93.21 & 60.74 \\
& RCL~\cite{RCL}~(TPAMI'23) & 44.94 & 74.43 & 82.99 & 92.31 & 71.93 & 86.84 & 92.96 & 73.18 \\
& RDE~\cite{RDE}~(CVPR'24) & 42.92 & 71.30 & 80.51 & 92.96 & 69.64 & 85.86 & 93.54 & 70.47 \\
& TME~\cite{TME}~(CVPR'25) & \underline{46.31} & \underline{75.78} & \underline{84.89} & \underline{95.83} & \underline{73.37} & \underline{88.02} & \underline{94.89} & \underline{74.58} \\
& \textbf{HABIT (Ours)} & \textbf{47.93} & \textbf{76.84} & \textbf{85.95} & \textbf{95.90} & \textbf{74.87} & \textbf{89.08} & \textbf{95.21} & \textbf{75.86} \\
\hline
\end{tabular}
\caption{Performance comparison on the CIRR test set in terms of R@K (\%) and R$_{\text{sub}}$@K (\%). The best and second-best results are highlighted in \textbf{bold} and \underline{underlined}, respectively.}
\label{tab:cirr-noise}
\end{table*}

\section{Experiments}
This section provides an in-depth examination, with $\sigma=0.2$ for ablation and sensitivity experiments, following TME.

\subsection{Experimental Settings}

\textbf{Datasets.}
Following previous works~\cite{TME, sprc}, we selected two widely adopted datasets for the CIR task, the fashion-domain dataset FashionIQ~\cite{FashionIQ}, and the open-domain dataset CIRR~\cite{cirr}.

\noindent
\textbf{Implementation Details.}
Our HABIT is trained with the learning rate of $5e\!\!-\!\!5$ with the AdamW optimizer on a V100 GPU. Following previous works~\cite{TME,sprc}, we utilize the pre-trained BLIP-2~\cite{blip-2} as the backbone of HABIT. The learnable query number $Q\!\!=\!\!32$. Regarding hyperparameter settings, we utilize the grid search to obtain the final value: $\kappa\!\!=\!\!10.0,\gamma\!\!=\!\!0.5$. The temperature coefficient $\tau=0.1$. Following TME~\cite{TME}, we introduce noise ratio $\sigma\!\!=\!\!\{0.0, 0.2, 0.5, 0.8\}$ during training to simulate NTC environment.

\noindent
\textbf{Evaluation.}
We adopt Recall@$K$ (R@$K$) as the primary metric. For CIRR, we report the performance of R@$\{1,5,10,50\}$ and further provide Recall$_\text{sub}$@$\{1,2,3\}$ metrics on its subset. For FashionIQ, we report the R@$\{10,50\}$ for each category (\textit{Dresses}, \textit{Shirts}, \textit{Tops\&Tees}). 

\subsection{Performance Comparison}
To assess the robustness and generalization of HABIT under the NTC scenario, we compare it with selected baselines: ordinary baselines (SSN, CALA, and SPRC) and robust baselines (RCL, RDE, and TME), on the CIRR and FashionIQ datasets across different noise ratios. As shown in Table~\ref{tab:fiq},\ref{tab:cirr-noise}, our analysis reveals the following:
\textbf{1)} HABIT outperforms existing robust methods in addressing NTC, demonstrating clear advantages in complex noise environments. On CIRR, HABIT achieves average improvements over TME of $1.16$\%, and $1.28$\% for $\sigma = 0.5$, and $0.8$, respectively. On FashionIQ, gains are $0.94$\%, and $1.31$\%. Notably, the performance gap with TME widens as noise increases. These gains result from HABIT's effective mining of mutual knowledge between multimodal queries and targets, precise noise estimation, and the integration of dual-consistency progressive learning.
\textbf{2)} Robust methods consistently surpass ordinary methods, with the performance gap widening as the noise ratio increases. For example, on the CIRR dataset at $\sigma =0.2$, the SOTA ordinary method SPRC even outperforms some robust models on certain metrics, and its Avg score is only $2.63$\% below HABIT. However, at $\sigma = 0.8$, SPRC's Avg score drops to $15.12$\% lower than HABIT, reflecting severe performance decline. This highlights the high sensitivity of traditional methods to noise and their instability in noisy correspondence scenarios, while robust models demonstrate superior noise resistance, emphasizing their necessity.

\subsection{Ablation Study}
To evaluate the efficacy of each HABIT module, we conduct comprehensive comparisons with variants in two groups:

\textbf{\textit{G1: Ablation on Mutual Knowledge Estimation.}}
This group examines the MKE module:
\textbf{D\#(1): w/o\_Sample} randomly selects samples instead of using standard samples Eq.(\ref{TR}) for mutual knowledge calculation.
\textbf{D\#(2): w/o\_TR} replaces the Transition Rate Eq.(\ref{estimation}) with the difference in mutual knowledge estimations.
\textbf{D\#(3): w/o\_MKE} removes the entire MKE process, setting all estimation values to $1$.
\textbf{\textit{G2: Dual-consistency Progressive Learning.}}
This group assesses DPL components:
\textbf{D\#(4): w/o\_CS} uses only current iteration estimations, omitting historical values in Eq.(\ref{noisymask}).
\textbf{D\#(5): w/o\_KL} removes the consistency constraint.
\textbf{D\#(6): w/o\_History} excludes both the consistency constraint and historical estimations.
\textbf{D\#(7): w/o\_mask} bypasses noisy mask computation, using only MKE estimations.
\textbf{D\#(8): w/o\_M\_Rank}, \textbf{D\#(9): w/o\_M\_Soft}, \textbf{D\#(10): w/o\_M\_KL} remove the noisy mask in Eq.(\ref{rank}), Eq.(\ref{soft}), and Eq.(\ref{kl}).
\textbf{D\#(11): w/o\_$\mathcal{L}_{rank}$} omits the robust contrastive loss.
\textbf{D\#(12): w/o\_$\mathcal{L}_{soft}$} omits the soft estimation margin loss.
\textbf{D\#(13): w/o\_$\mathcal{L}_{KL}\&\mathcal{L}_{soft}$} removes both consistency constraint and soft estimation margin loss.

\begin{table}[h]
  \centering
  \small
    \tabcolsep=5pt
    \begin{tabular}{c|l|cc|cc}
    \hline
    \multicolumn{1}{c|}{\multirow{2}{*}{D\#}}
    &\multicolumn{1}{c|}{\multirow{2}{*}{Deriv.}} 
    & \multicolumn{2}{c|}{FashionIQ-Avg.} 
    & \multicolumn{2}{c}{CIRR-Avg.} \\
    \cline{3-6}
    & & R@10 & R@50 
    & R@K & R$_{\text{sub}}$@K \\
    \hline\hline
    \multicolumn{6}{c}{\textbf{\textit{(a) Mutual Knowledge Estimation (MKE)}}} \\
    \hline
    1 & w/o Sample      & 53.77 & 73.01 & 79.42 & 88.12 \\
    2 & w/o TR        & 53.40 & 73.03 & 79.58 & 87.71 \\
    3 & w/o MKE      & 53.08 & 73.12 & 79.32 & 87.94 \\
    \hline
    \multicolumn{6}{c}{\textbf{\textit{(b) Dual-consistency Progressive Learning (DPL)}}} \\
    \hline
    4 & w/o CS     & 53.30 & 73.22 & 79.48 & 88.00 \\
    5 & w/o KL         & 53.44 & 73.92 & 79.27 & 87.28 \\
    6 & w/o History  & 53.66 & 73.87 & 78.32 & 87.24 \\
    7& w/o mask      & 53.59 & 73.67 & 79.46 & 87.72 \\
    8 & w/o M\_Rank & 53.48 & 73.58 & 79.67 & 87.98 \\
    9 & w/o M\_Soft      & 53.51 & 73.65 & 79.46 & 88.00 \\
    10 & w/o M\_KL      & 53.73 & 73.47 & 79.71 & 88.03 \\
    11 & w/o $\mathcal{L}_{rank}$    &  51.03 &  71.82 & 77.92 & 87.15 \\
    12 & w/o $\mathcal{L}_{soft}$         & 53.14 & 73.34 & 79.38 & 88.21 \\
    13 & w/o $\mathcal{L}_{KL}\&\mathcal{L}_{soft}$     & 53.27 & 73.69 & 79.51 & 88.20 \\
    \hline
    \multicolumn{2}{c|}{\textbf{HABIT~(Ours)} }
    & \textbf{54.48} & \textbf{74.28} 
    & \textbf{79.94} & \textbf{88.87} \\
    \hline
    \end{tabular}
  \caption{Ablation study on FashionIQ and CIRR datasets.}
  \label{tab:abla}
\end{table}

Key findings from Table~\ref{tab:abla} include:
\textbf{1)} Removing $\mathcal{L}_{rank}$ \textbf{(D\#(11))} yields the worst performance, underscoring the essential role of robust contrastive loss in noise suppression and retrieval accuracy.
\textbf{2)} Omitting any component of the \textbf{MKE} module leads to performance degradation. Notably, eliminating standard samples for mutual knowledge \textbf{(D\#(1))} degrades performance, though less so than removing all MKE estimations \textbf{(D\#(3))}. Replacing the transition rate with mutual knowledge discrepancy \textbf{(D\#(2))} results in the largest drop, highlighting the transition rate's importance for capturing intrinsic NTC relationships.
\textbf{3)} Removing either the historical estimation sequence \textbf{(D\#(4))} or the ``calibrate bad habits'' constraint \textbf{(D\#(5))} reduces performance; removing both \textbf{(D\#(6))} causes even greater decline. These help retain good habits and noise discrimination, preventing semantic inconsistency over time-flux training.
\textbf{4)} Excluding the noisy mask \textbf{(D\#(7))} decreases performance, validating its role in filtering noisy triplets. Further, removing the noisy mask from any loss function \textbf{(D\#(8)-D\#(10))} results in performance drops, reflecting its broad optimization effect via knowledge aggregation.
\textbf{5)} Excluding $\mathcal{L}_{soft}$ \textbf{(D\#(11), D\#(12))} impairs performance, indicating that the soft estimation margin loss reduces overfitting from partial matches and misclassified noisy samples.

\subsection{Case Study}
\begin{figure}[h]
\begin{center}
\includegraphics[width=\linewidth]{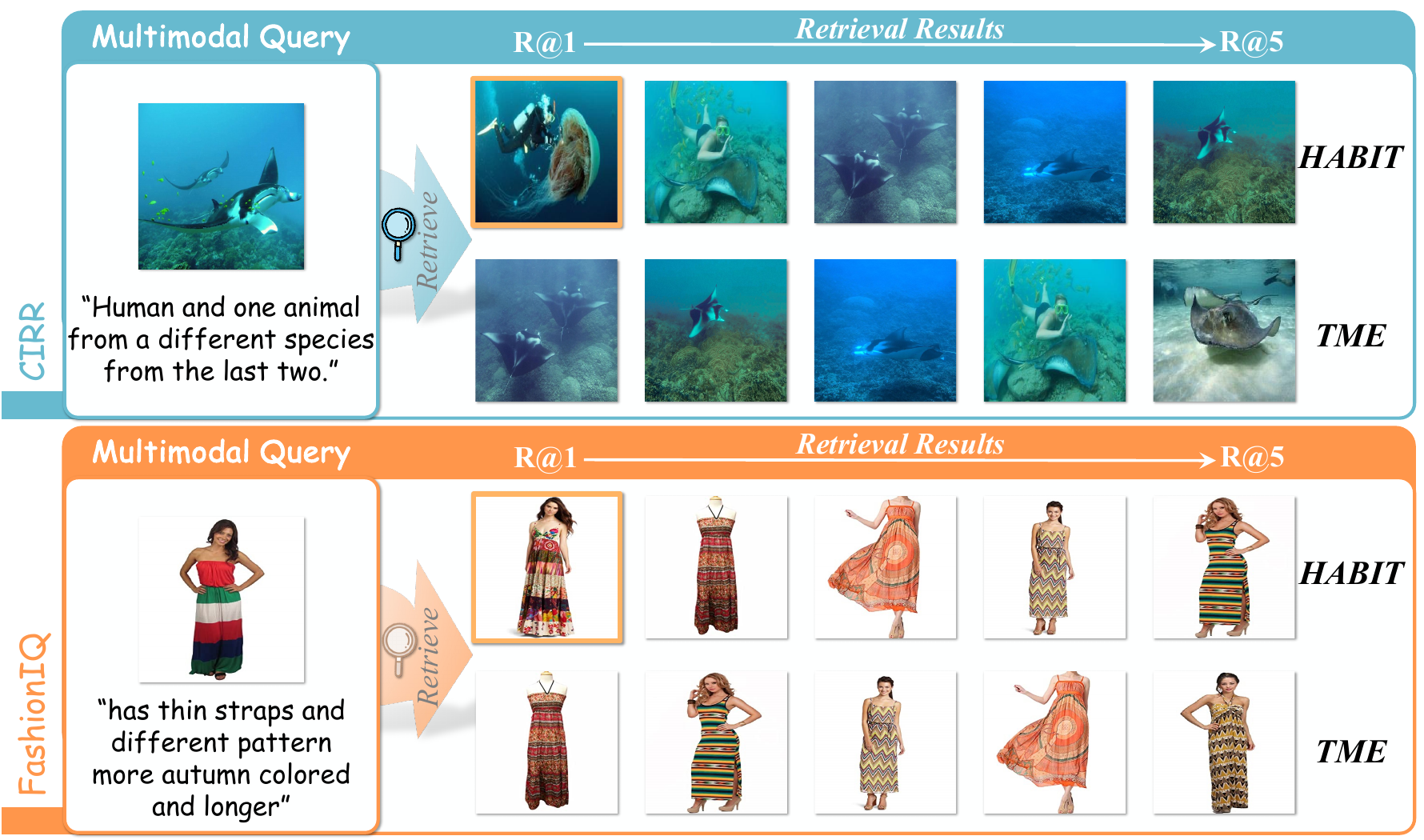}
\end{center}
   \caption{Case Study on CIRR and FashionIQ.}
\label{fig:lrq}
\end{figure}

Figure~\ref{fig:lrq} presents the top-$5$ results from HABIT and the SOTA robust CIR model TME on two CIR datasets. In the CIRR example (Figure~\ref{fig:lrq}(a)), HABIT correctly retrieves the top-ranked image featuring both a diver and a sea turtle, fully satisfying the compositional semantics. By contrast, TME fails to interpret the cross-entity relationship, retrieving only manta ray images and missing the ``human+different species'' requirement, revealing its limited grasp of fine-grained semantic composition.
In Figure~\ref{fig:lrq}(b), HABIT's top-$1$ image matches all target attributes (e.g., reddish-brown long dress, warm autumn tones) at top-$1$, with subsequent results also aligning well with key attributes. In comparison, TME's top-$1$ result, although color-similar, lacks thin straps.
Overall, HABIT reveals clear advantages in handling multi-level semantic compositions, which stem from its meticulous understanding of triplet semantic relationships. 

\section{Conclusion}
In this study, we investigated the NTC problem in the CIR task. To address the challenges of the precise estimation of composed semantic discrepancy and the insufficient progressive adaptation to modification discrepancy, we proposed HABIT, which comprised two key modules. The MKE module, employed the transition rate of variational mutual information to achieve accurate noise-aware label assignment. Furthermore, the DPL module introduced a collaborative mechanism between the historical and current models, progressively enhancing the model's understanding of the complex semantic relations within triplets and continuously reduced the probability of misdetermination. Extensive experiments showed that HABIT outperformed most methods under various noise levels, demonstrating its superiority and robustness.

\section*{Acknowledgments}
This work was supported in part by the National Natural Science Foundation of China, No.:62276155, No.:62576195, and No.:62572282; in part by the China National University Student Innovation \& Entrepreneurship Development Program, No.:2025282 and No.:2025283.

\clearpage
\appendix
\noindent
This is the supplementary material of the submitted paper \textit{\textbf{``HABIT: Chrono-Synergia Robust Progressive Learning Framework for Composed Image Retrieval''}}. The content catalog is as follows:
\begin{itemize}
    \item \textbf{Appendix~\ref{appendix:datasets}}: Datasets
        \item \textbf{Appendix~\ref{appendix:additional_performance_cirr_fiq}}: Comprehensive Performance Comparison on CIRR and FashionIQ
        \item \textbf{Appendix~\ref{appendix:additional_performance_efficiency}}: Efficiency Evaluation

    \item \textbf{Appendix~\ref{appendix:code}}: HABIT Training Procedure and Dynamic Margin Calculation
    \begin{itemize}
        \item \textbf{Appendix~\ref{appendix:code_algorithm}}: Algorithm of HABIT's Training Procedure
        \item \textbf{Appendix~\ref{appendix:code_calculation}}: The Calculation of Dynamic Margin
    \end{itemize}
    \item \textbf{Appendix~\ref{appendix:qualitative}}: More Qualitative Results
    \begin{itemize}
     \item \textbf{Appendix~\ref{appendix:qualitative_more_case}}: Failure Cases Analysis
        \item \textbf{Appendix~\ref{appendix:noisy_identify}}: NTC Recognition Situation
         \item \textbf{Appendix~\ref{appendix:qualitative_matrix}}: Similarity Matrix

    \end{itemize}
\end{itemize}

\section{Datasets}
\label{appendix:datasets}
To thoroughly assess the performance of our proposed model, HABIT, we utilized two widely recognized benchmark datasets: FashionIQ, which focuses on the fashion domain, and CIRR, which targets open-domain scenarios. A detailed description of each dataset is presented below.
\begin{itemize}
\item \textbf{FashionIQ}~\cite{FashionIQ} serves as a benchmark dataset tailored for fashion-focused composed image retrieval. It comprises $77,684$ web-crawled fashion photos, organized into $30,134$ annotated triplets spanning three key clothing categories: \textit{dresses}, \textit{shirts}, and \textit{tops\&tees}. This dataset is primarily used to assess the retrieval capability of models based on the alignment between visual content and descriptive textual modifications in the fashion context.

\item \textbf{CIRR}~\cite{cirr} is developed using real-world images sourced from the NLVR2 dataset~\cite{NLVR2}, which focuses on natural language visual reasoning. It includes $36,554$ annotated triplets and a total of $21,552$ unique images. In contrast to FashionIQ, CIRR places greater emphasis on complex multi-object interactions within diverse natural environments, reducing the risk of overfitting to a narrow domain. Furthermore, it addresses the issue of annotation sparsity, which is commonly seen in datasets like FashionIQ that may produce many false negatives, by incorporating a dedicated subset designed for fine-grained contrastive testing. CIRR is ideal for assessing a model's ability to process intricate scenes that require nuanced multi-modal understanding.
\end{itemize}

\begin{algorithm}[h!]
\caption{The HABIT's Training Procedure}
\label{alg:habit}
\textbf{Input}: Noisy triplet set $\mathcal{T} = \{ \langle x_r, x_m, x_t \rangle \}$\\
\textbf{Parameter}: Max epochs $N$, batch size $B$, temperature $\tau$\\
\textbf{Output}:Trained model parameters
\begin{algorithmic}[1] 
\FOR{iteration $I = 1$ to $N$}
    \STATE Sample a batch $\{x_r, x_m, x_t\}_{b=1}^{B} \subset \mathcal{T}$
    
    \STATE // Feature Extraction via Q-Former
    \STATE $\mathbf{F}_c \gets \text{Q-Former}(\varPhi_{\mathbb{I}}(x_r), \varPhi_{\mathbb{T}}(x_m))$
    \STATE $\mathbf{F}_t \gets \text{Q-Former}(\varPhi_{\mathbb{I}}(x_t))$
    
    \STATE // Mutual Knowledge Estimation (MKE)
    \STATE $\text{MK} \gets \operatorname{MutualInfo}(\mathbf{F}_c, \mathbf{F}_t)$
    \STATE $\text{MK}^s \gets \min(\text{MK})$ \COMMENT{Standard sample}
    \STATE Compute transition rates: $\text{TR}_{c,t}$, $\text{TR}_{c,t^s}$, $\text{TR}_{t,c^s}$
    \STATE $\mathbb{E}_b \gets \left(1 + \text{TR}_{c,t} + |\text{TR}_{c,t^s} - \text{TR}_{t,c^s}|\right)^{-1}$

    \STATE // Chrono-Synergia Noise Discrimination
    \STATE $\mathbf{e}^{(I)} \gets [\mathbb{E}_1, ..., \mathbb{E}_B]$
    \STATE $\mathcal{O}^{(I)} \gets \text{DBSCAN}(\mathbf{e}^{(I)})$
    \STATE $\mathcal{O}^{(I-1)} \gets \text{DBSCAN}(\mathbf{e}^{(I-1)})$
    \FOR{$b = 1$ to $B$}
        \IF{$b \in \mathcal{O}^{(I)} \cap \mathcal{O}^{(I-1)}$}
            \STATE $m_b \gets 0$ \COMMENT{Noisy sample}
        \ELSE
            \STATE $m_b \gets 1$ \COMMENT{Clean sample}
        \ENDIF
    \ENDFOR

    \STATE // Similarity Computation
    \STATE $\mathbf{s}^{(I)} \gets \operatorname{SimilarityMatrix}(\mathbf{F}_c, \mathbf{F}_t)$

    \STATE // Time-Flux Knowledge Updating
    \STATE $\mathcal{L}_{\text{KL}} \gets \frac{1}{B} \sum\limits_{b=1}^{B} D_{\text{KL}} \left( \mathbf{s}_b^{(I)} \cdot m_b \,\Vert\, \mathbf{s}_b^{(I-1)} \cdot m_b \right)$

    \STATE {\small $\mathcal{L}_{\text{soft}} \!\gets\! \!\frac{1}{B}\! \sum\limits_{b=1}^{B} \left( \max_{j \neq b} \left[ \text{margin}(\mathbb{E}_b) + s_{bj}^{(I)} \!-\! s_{bb}^{(I)} \right]_+ \!\cdot\! m_b \right)$}
    
    \STATE {\small $\mathcal{L}_{\text{rank}} \gets \frac{1}{B} \sum\limits_{b=1}^{B} -\log \left( \left(1 - \operatorname{Softmax}(\mathbf{s}_b^{(I)} / \tau)\right) \cdot m_b \right)$}

    \STATE $\mathcal{L}_{\text{HABIT}} \gets \mathcal{L}_{\text{KL}} + \mathcal{L}_{\text{soft}} + \mathcal{L}_{\text{rank}}$
    \STATE Update model using $\mathcal{L}_{\text{HABIT}}$
\ENDFOR
\STATE \textbf{return} Trained model parameters
\end{algorithmic}
\end{algorithm}

\section{Comprehensive Performance Comparison on CIRR and FashionIQ}
\label{appendix:additional_performance_cirr_fiq}
\begin{figure*}[h]
\begin{center}
\includegraphics[width=0.95\linewidth]{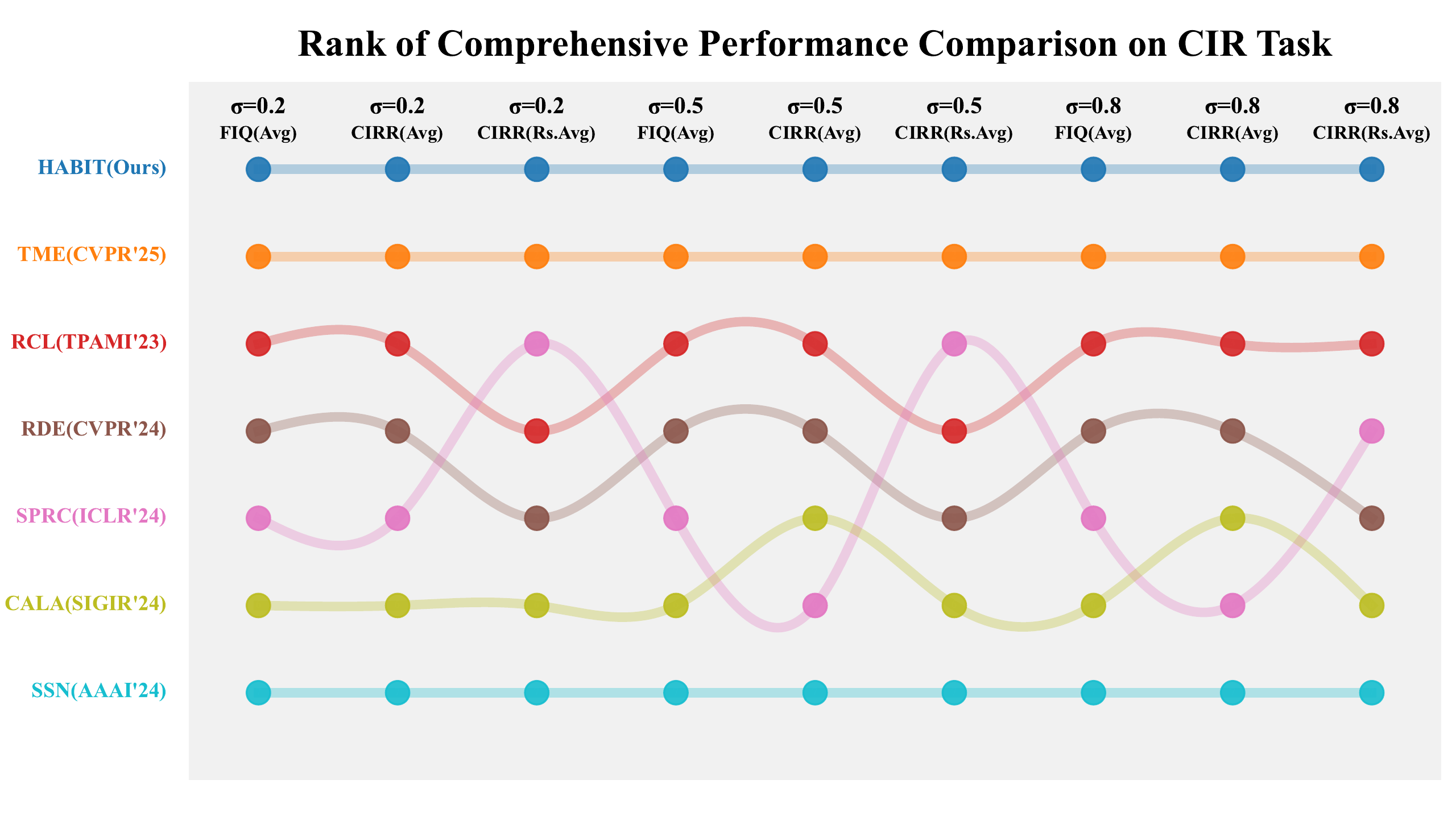}
\end{center}
   \caption{Comprehensive Performance Comparison Rank on CIRR and FashionIQ.}
\label{fig:rank}
\end{figure*}

\begin{table*}[ht]
  \tabcolsep=1.0pt
  \centering
    \begin{tabular}{c|c|c|c|c|c|c|c|c}
    \hline
    Type  & Method & FLOPs & Parameters & GPU Memory & Test time & Train Time & FIQ-Avg & CIRR-Avg \\
    \hline
    \hline
    Ordinary & SPRC  & 413.38G & 915.69 M & 24.48GiB & 0.011s/sample & {2.62s/iteration} & 56.33 & 76.98 \\
    \hline
    \multirow{4}{*}{Robust Methods} & TME   & 405.2G & {915.68M} & 12.41GiB & 0.124s/sample & 7.86s/iteration & 63.97 & 79.74 \\
    \cline{2-9}
    & w/o History & {393.08G} &915.69M& {11.37GiB} & {0.0089s/sample} & 2.85s/iteration & 63.76 &  78.67  \\
    & {HABIT (Ours)} & {393.08G} &915.69M & {11.37GiB} & {0.0089s/sample} &  2.94s/iteration & {64.38} & {79.61} \\
    \hline
    \end{tabular}%
\caption{Comparison of computational complexity and efficiency among SPRC, TME, and HABIT with its ablation variants.}
  \label{tab:efficiency}%
\end{table*}%

Figure~\ref{fig:rank} presents a comparative analysis of different models under varying noise ratios ($\sigma$). In this figure, CIRR Avg denotes the average value of R@K on the CIRR dataset, CIRR Rs.Avg refers to the average of R$_{\text{sub}}$@K on the CIRR dataset, and FIQ(Avg) represents the mean R@K across the three subsets of the FashionIQ dataset. Each row corresponds to model performance under a specific noise ratio ($\sigma = 0.2, 0.5, 0.8$), where colors indicate different models and the lines illustrate the variation in performance across noise levels. This figure clearly reveals the relative performance of each model under different levels of noise. The results demonstrate that HABIT consistently achieves the best performance across all datasets and noise conditions. This robustness can be attributed to the accurate noise-aware estimation based on the Transition Rate of mutual information and the integration of Dual-consistency Progressive Learning. These components enable HABIT to effectively adapt to noise interference and exhibit strong generalization capabilities in Composed Image Retrieval tasks. In contrast, other models such as RDE and SPRC exhibit considerable fluctuations in performance across different noise ratios and datasets, indicating limited stability in the presence of noise and complex scenarios. Although the TME model maintains relatively stable performance under varying noise and dataset conditions, it consistently ranks second and demonstrates lower recall rates than HABIT. This suggests that TME possesses limited robustness against noise interference and lacks the adaptability of HABIT in challenging environments.

\begin{figure*}[ht]
\begin{center}
\includegraphics[width=0.8\linewidth]{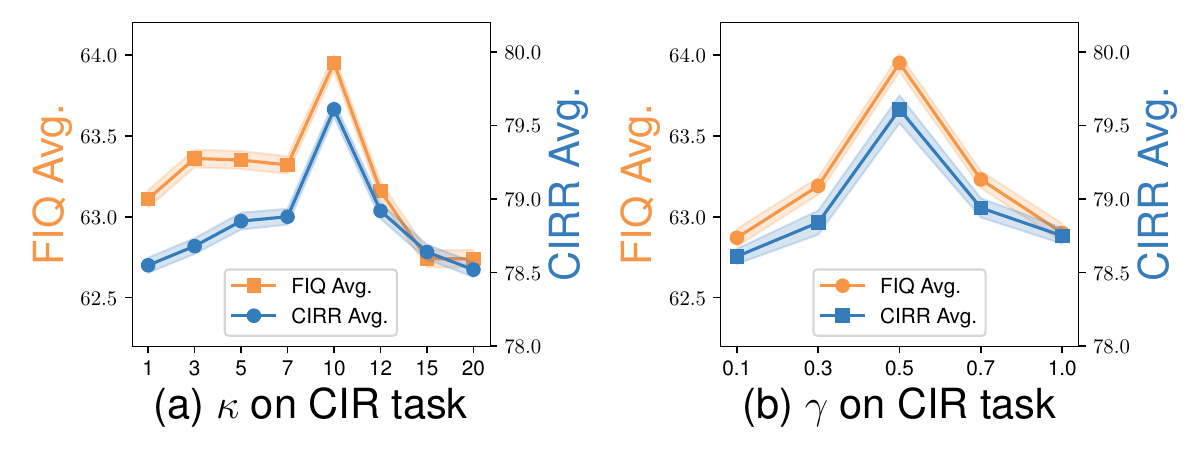}
\end{center}
   \caption{Sensitivity to the hyperparameters (a) $\kappa$ and (b) $\gamma$.}
\label{fig:sensi}
\end{figure*}

\section{Efficiency Evaluation}
\label{appendix:additional_performance_efficiency}
To evaluate system efficiency and resource consumption, we compare the ordinary baseline (SPRC), the robust baseline (TME), and HABIT, along with its variant without historical consistency, under identical hardware conditions and a batch size of 128. The following findings are derived from Table~\ref{tab:efficiency}:

(1) In terms of \textbf{computational cost and memory usage}, HABIT achieves 393.08G FLOPs, representing a reduction of approximately 2.99\% compared to TME (405.2G) and 4.91\% relative to SPRC (413.38G). Its memory consumption decreases from 12.41GiB (TME) to 11.37GiB (a reduction of about 8.38\%) and is further reduced by approximately 53.55\% compared to SPRC's 24.48GiB. All models have the same parameter count of 915.69M, indicating that HABIT's performance improvements are not attributed to additional parameters.

(2) Regarding \textbf{speed}, HABIT achieves a testing time of 0.0089 seconds per sample, offering approximately 13.93 times acceleration compared to TME (0.124 seconds/sample), and around 1.24 times speedup over SPRC (0.011 seconds/sample). Its training time significantly drops from 7.86 seconds per iteration (TME) to 2.94 seconds per iteration (a reduction of approximately 62.60\%). Although slightly slower than SPRC (2.62 seconds per iteration, approximately 12.21\% difference), SPRC requires substantially more memory (24.48GiB), which limits its practical deployment and scalability.

(3) In terms of \textbf{overall performance considering both speed and accuracy}, HABIT achieves higher retrieval precision, with FIQ-Avg = 64.38 and CIRR-Avg = 79.61. This performance improvement strikes an excellent balance with the efficiency overhead.

(4) From the \textbf{ablation perspective}, the w/o History variant exhibits identical FLOPs, memory consumption, and testing time as HABIT, indicating that the Dual-consistency Progressive Learning (DPL) module operates solely during training and introduces negligible inference overhead. Additionally, HABIT increases training time by only approximately 3.16\% compared to w/o History (2.94s vs. 2.85s), while yielding notable performance gains on CIRR and FIQ.

(5) From the \textbf{architectural perspective}, HABIT incorporates Mutual Knowledge Estimation (MKE) and DPL to perform intra-batch mutual knowledge estimation and Dual-consistency Progressive Learning. These modules introduce only lightweight computation during training, without adding inference branches or re-ranking steps. As a result, HABIT maintains the same parameter size while simultaneously reducing computational cost, memory usage, and latency compared to TME and SPRC. The minor training overhead is effectively translated into stable performance gains and enhanced employability.

\subsection{Sensitivity Analysis}

To assess HABIT's sensitivity to hyperparameters $\kappa$ and $\gamma$, we analyze performance on CIRR and FashionIQ, as shown in Figure~\ref{fig:sensi}. In subfigure~(a), increasing $\kappa$ (which weights $\mathcal{L}_{KL}$) from $1$ to $20$ initially improves performance, then leads to a decline. This suggests that moderate collaboration between current and historical model knowledge helps preserve learned ``good habits'' and matching consistency, while excessive weighting makes the model overly dependent on historical similarities, hindering adaptation and limiting knowledge acquisition.
In subfigure~(b), $\gamma$, which controls $\mathcal{L}_{soft}$, shows a similar trend: as $\gamma$ rises from $0.1$ to $1.0$, performance first improves, then slightly drops. This reflects the dual role of soft estimation margin loss, dynamically penalizing hard negatives to calibrate bad habits and gains noise robustness, but an excessively large $\gamma$ can cause optimization bias and interfere with semantic alignment.

\section{HABIT Training Procedure and Dynamic Margin Calculation}
\label{appendix:code}
\subsection{Algorithm of HABIT's Training Procedure}
\label{appendix:code_algorithm}
To support the main methodology, we present the complete training workflow of HABIT as pseudocode in Algorithm~\ref{alg:habit}. This representation clearly and reproducibly outlines how the MKE and DPL components are collaboratively optimized throughout the training process.

\subsection{The Calculation of Dynamic Margin}
\label{appendix:code_calculation}
In this section, we detail the calculation of the $\operatorname{margin}(\cdot)$ used in the soft estimation margin loss. Specifically, following~\cite{noisy-nips}, the calculation process is a dynamic margin that depends on the estimation value $\textbf{e}_b^{(I)}$ of the b-th sample in the batch, formulated as,
\begin{equation}
    \operatorname{margin}(\textbf{e}_b^{(I)})=m \times (10^{\textbf{e}_b^{(I)}}-1) / 9
\end{equation}
where $m$ denotes the margin parameter, which is empirically set to $0.2$, following~\cite{noisy-nips}.

\section{More Qualitative Results}
\label{appendix:qualitative}

\subsection{Failure Cases Analysis}
\label{appendix:qualitative_more_case}

\begin{figure*}[h]
\begin{center}
\includegraphics[width=0.95\linewidth]{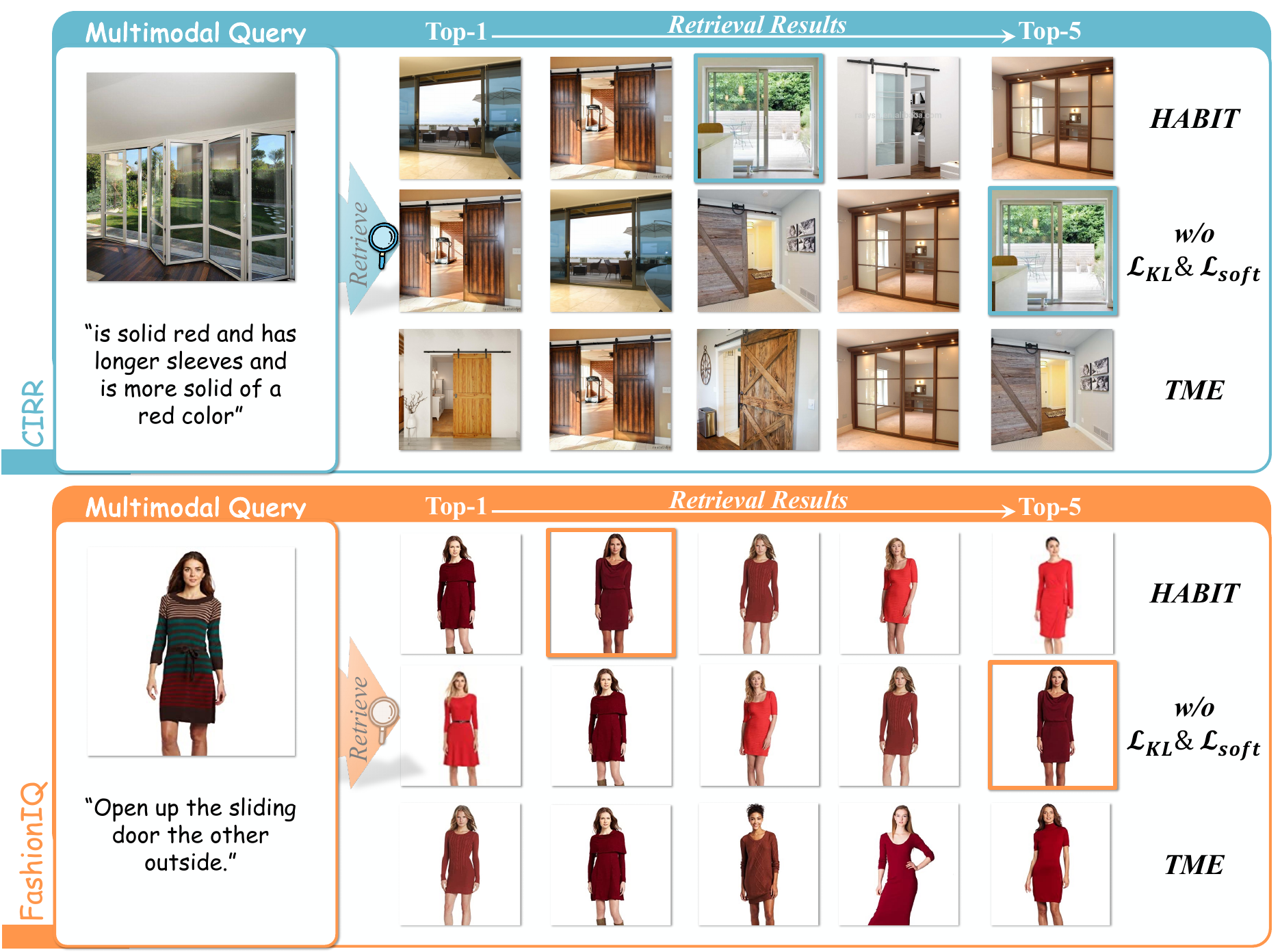}
\end{center}
   \caption{Failure Cases on CIRR and FashionIQ datasets.}
\label{fig:failure_case}
\end{figure*}

In real-world CIR tasks, even models equipped with robust learning strategies, such as HABIT, inevitably encounter challenging failure cases. 
Figure~\ref{fig:failure_case} presents representative examples from both the CIRR and FashionIQ datasets, allowing for a closer examination of how the model responds to complex semantic transformations.

In the CIRR failure case, the multimodal query is “is solid red and has longer sleeves and is more solid of a red color,” with a reference image depicting a glass sliding door. 
This represents a complex scene modification requiring the model to reason over both composition semantics and visual changes. 
HABIT retrieves a top-1 result that, although not annotated as the ground-truth, closely matches the query in terms of color and structure. 
Most of its top candidates also satisfy the required details in the query, reflecting HABIT's strong ability to model multi-level semantic changes and capture subtle visual cues. In contrast, the ablated variant w/o $L_{KL}$ \& $L_{soft}$ shows a noticeable drop in accuracy, with retrieval results exhibiting scene drift and partial mismatches, demonstrating the importance of these two components in preserving beneficial learning habits during training. 
TME, meanwhile, is more prone to semantic ambiguity and tends to return candidates with greater divergence from the true target in terms of fine-grained semantics. 
This comparison illustrates that HABIT, with its dual-consistency progressive learning mechanism, can better capture and adapt to complex changes. Even when the annotated ground-truth is not retrieved, the outputs remain highly relevant to the user intent, highlighting the practical value of the model's semantic understanding.

For the FashionIQ failure case, the query is “Open up the sliding door the other outside,” with a reference image of a red long-sleeved dress. 
While HABIT does not rank the ground-truth target first, all of its top-5 retrievals are highly similar in style, color, and type, and some candidates may even better fit the user's real-world intent than the labeled ground-truth. The ablation model w/o $L_{KL}$ \& $L_{soft}$ shows degraded performance, with a clear drop in relevance among retrieved samples. 
TME, meanwhile, retrieves even more candidates that deviate from the intended query semantics. Upon further analysis, we find that some of these failure cases actually reflect the presence of false negatives in the dataset: retrieved results are visually and semantically correct according to the query, yet not marked as targets due to annotation limitations.

Overall, HABIT's consistently high Top-k relevance in these failure cases is attributed to its accurate triplet correspondence modeling and advanced semantic reasoning capabilities. Especially under complex transformation, fine-grained attribute matching, and implicit query scenarios, HABIT effectively fuses reference image and modification text to better align composed features with the true user need. 
This not only validates the contribution of mutual knowledge estimation and dual-consistency progressive learning to model robustness, but also reveals new challenges for annotation and evaluation in current CIR benchmarks.

\subsection{NTC Recognition Situation}
\label{appendix:noisy_identify}

\begin{figure*}[h]
\begin{center}
\includegraphics[width=0.95\linewidth]{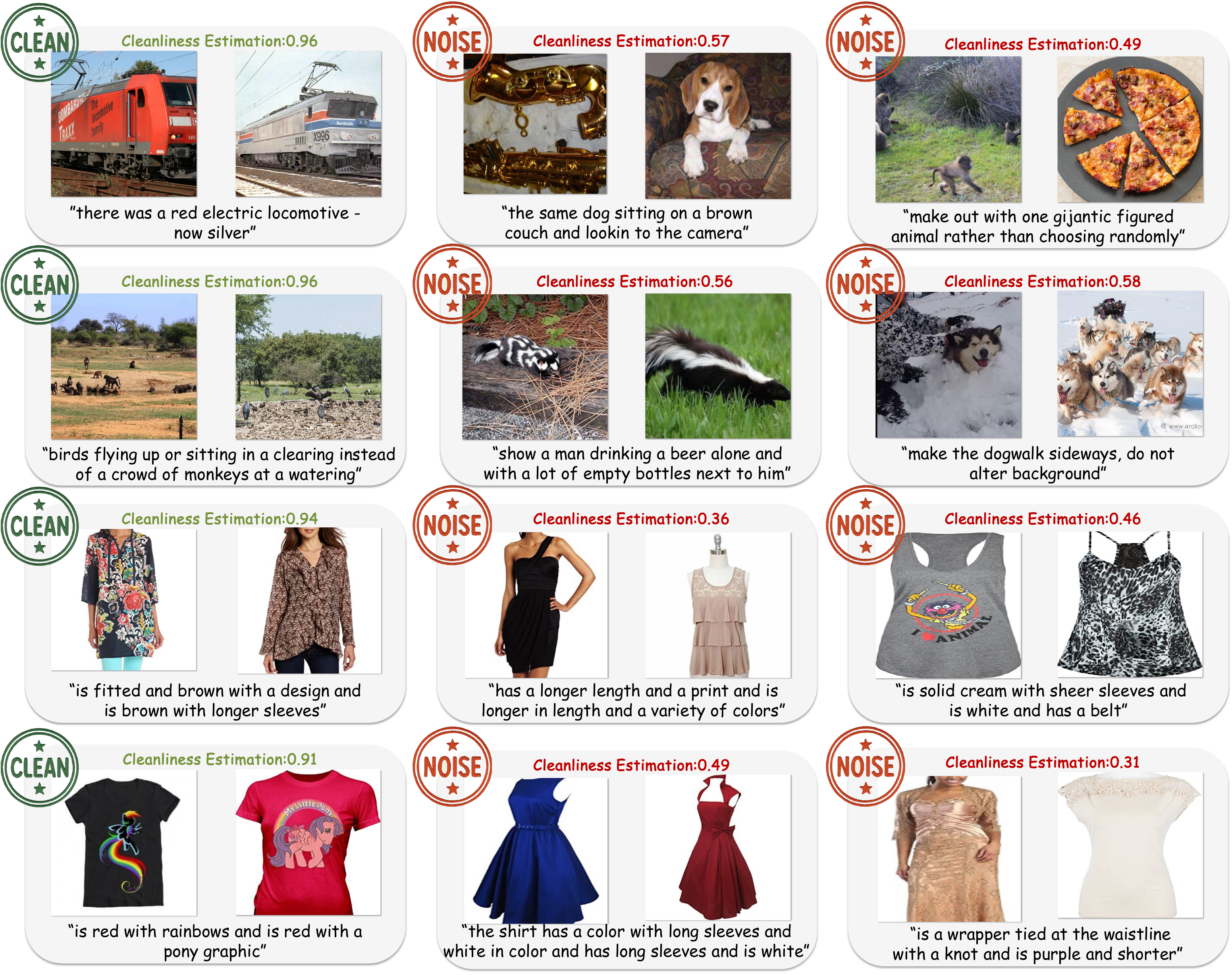}
\end{center}
   \caption{Visualization of the cleanliness estimation.}
\label{fig:ntc}
\end{figure*}

As shown in Figure~\ref{fig:ntc}, we present the model's clean/noise discrimination on CIR triplets under NTC scenarios, along with the estimated cleanliness scores calculated by the MKE module. 
The results demonstrate that the model is able to accurately distinguish between clean and noisy matches, outputting reasonable cleanliness estimates and thereby exhibiting strong discriminative capability under Noise Triplet Correspondence (NTC) conditions.

Specifically, the green area on the left highlights triplets identified as clean by the model, which are generally assigned high confidence scores. For example, in the top-left of the first row, the reference and target images are a red electric locomotive and a silver locomotive, with the query text ``there was a red electric locomotive – now silver.” The model assigns a cleanliness estimation of $0.96$, accurately reflecting a single, significant attribute change between the images. Similarly, in the second row, the triplet ``birds flying up or sitting in a clearing instead of a crowd of monkeys at a watering” also receives a high confidence score of $0.96$, indicating the model's ability to capture subtle semantic changes in complex multi-object scenes.

In contrast, the red area on the right displays noisy correspondence, which are given noticeably lower confidence. For instance, the top-right example in the first row has the text ``make out with one gigantic figured animal rather than choosing randomly,” while the target image is a pizza and an animal, a clear case of cross-category (semantically irrelevant) mismatch. The model correctly classifies this as noise, assigning a cleanliness estimation of only $0.49$. Similarly, in the third row, for the description ``the shirt has a color with long sleeves and is white in color and has long sleeves and is white,” the blue and white dresses are mismatched, and the model outputs a low cleanliness score of $0.49$, showing its sensitivity to detailed discrepancies between images and textual descriptions.

Additionally, some boundary cases that are difficult to distinguish receive appropriately intermediate confidence scores. For example, on the right of the second row, the composition of a dog and a pack of wolves with the description ``make the dogwalk sideways, do not alter background” is visually ambiguous, yet the model assigns a moderate confidence of $0.58$, demonstrating its reasonable handling of challenging scenarios.

Overall, this visualization strongly supports the model's capability for accurate cleanliness estimation of triplets in NTC settings. The model not only precisely identifies clean semantic correspondences but also reliably assigns low cleanliness scores to noisy matches, facilitating robust downstream training and dynamic label correction. 
Such fine-grained, confidence-based discrimination effectively mitigates the impact of noisy samples in CIR tasks, significantly enhancing the model's practical robustness and generalization performance.

\subsection{Similarity Matrix}
\label{appendix:qualitative_matrix}

\begin{figure}[h]
\begin{center}
\includegraphics[width=0.95\linewidth]{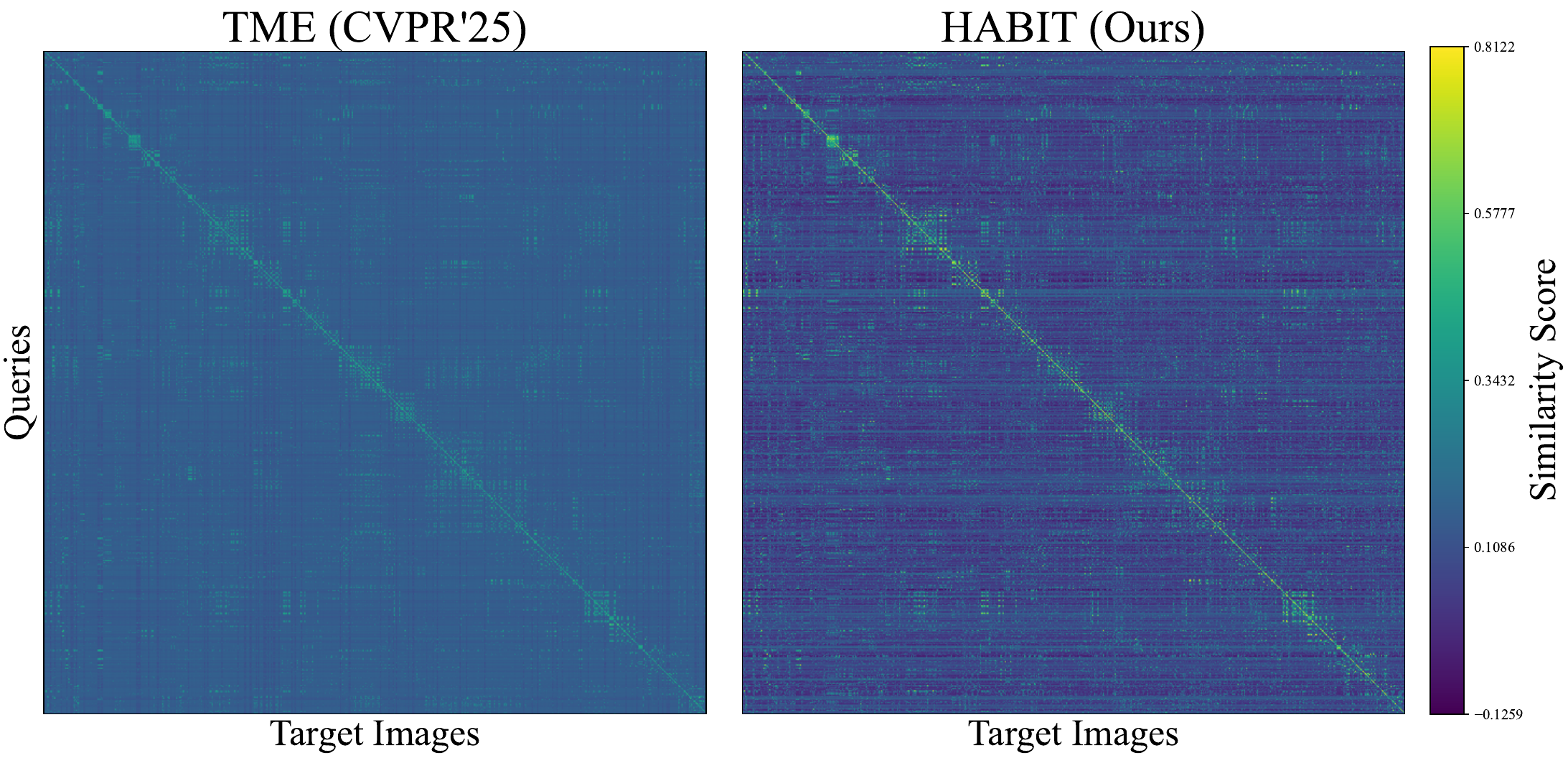}
\end{center}
   \caption{Similarity Matrix of TME and HABIT on CIRR dataset.}
\label{fig:sim_mat}
\end{figure}

As shown in Figure~\ref{fig:sim_mat}, we compare the similarity matrices of TME (CVPR'25)~\cite{TME} and HABIT (Ours) on the CIRR dataset, where the horizontal axis represents target images and the vertical axis denotes multimodal queries. A higher similarity score indicates a stronger model belief that the target matches the given query.

From the figure, we observe the following phenomena and their causes: 
\textbf{1) Brighter and sharper diagonal.} The diagonal entries of HABIT are notably higher than those of TME, indicating that the similarity of positive pairs (the query and its target image) is further increased; the positive peaks are more concentrated, making top-ranked hits easier to obtain. 
\textbf{2) Darker off-diagonal.} In HABIT, regions far from the diagonal are visibly darker, showing that similarities for negatives are systematically suppressed and spurious high-similarity negatives are effectively reduced. 
\textbf{3) Larger positive–negative gap.} The contrast between the diagonal and off-diagonal is higher with HABIT, reflecting a larger margin between positives and negatives, which is consistent with the improvements on CIRR in R@1/5/10 and $R_{\text{sub}}@K$. 

These differences stem from two aspects: first, Mutual Knowledge Estimation (MKE) anchors on a Standard Sample and uses the Transition Rate of mutual knowledge as a relative metric, producing more reliable cleanliness estimation and down-weighting noisy correspondence and hard negatives; second, Dual-consistency Progressive Learning (DPL) combines Chrono-Synergia noise discrimination, KL-based consistency, and the soft estimation margin loss to calibrate the model along the Time-Flux, push negatives away, and preserve correct determination, which manifests as a brighter diagonal and a darker off-diagonal in the similarity matrix.

\small
\bibliography{aaai2026}

\end{document}